\documentclass[conference]{IEEEtran}
\IEEEoverridecommandlockouts
\usepackage{cite}
\usepackage{amsmath,amssymb,amsfonts}
\usepackage{algorithmic}
\usepackage{graphicx}
\usepackage{textcomp}
\usepackage{xcolor}
\usepackage{bbding}
\usepackage{pifont}
\usepackage{wasysym}
\usepackage{amssymb}

\def\BibTeX{{\rm B\kern-.05em{\sc i\kern-.025em b}\kern-.08em
    T\kern-.1667em\lower.7ex\hbox{E}\kern-.125emX}}
    
\usepackage{tikz}
\newcommand*\circled[1]{\tikz[baseline=(char.base)]{
    \node[shape=circle,draw,inner sep=1pt] (char) {#1};}}
\begin{document}

\title{Challenges and Obstacles Towards Deploying \\Deep Learning Models on Mobile Devices}

\author{\IEEEauthorblockN{Hamid Tabani*, Ajay Balasubramaniam*\thanks{$*$ Equal Contribution.}, Elahe Arani, Bahram Zonooz}
\IEEEauthorblockA{Advanced Research Lab, NavInfo Europe, Eindhoven, The Netherlands\\
\texttt{\{hamid.tabani, ajay.balasubramaniam, elahe.arani,\}@navinfo.eu, bahram.zonooz@gmail.com}}
}
\maketitle

\begin{abstract}
From computer vision and speech recognition to forecasting trajectories in autonomous vehicles, deep learning approaches are at the forefront of so many domains. Deep learning models are developed using plethora of high-level, generic frameworks and libraries. Running those models on the mobile devices require hardware-aware optimizations and in most cases converting the models to other formats or using a third-party framework. 
In reality, most of the developed models need to undergo a process of conversion, adaptation, and, in some cases, full retraining to match the requirements and features of the framework that is deploying the model on the target platform.
Variety of hardware platforms with heterogeneous computing elements, from wearable devices to high-performance GPU clusters are used to run deep learning models. In this paper, we present the existing challenges, obstacles, and practical solutions towards deploying deep learning models on mobile devices. 
\end{abstract}


\section{Introduction}
\label{sec:introduction}

Deep learning is at the forefront of a broad range of tasks including but not limited to image segmentation and object detection. Deep learning methods have far high accuracy than any other alternative solution, in many situations, and in some domains, they are the only solution possible.
With the broad popularity of these approaches, the need for accelerated AI development with a seamless transition from prototyping to development and deployment is inevitable. To this end, deep learning frameworks offer high-level and streamlined programming interfaces which provide access to building blocks for designing, training, validating, and analyzing deep learning models.
Many deep learning frameworks have climbed to become the latest choice of academics and industry professionals since deep learning regained popularity in 2012. 

This deluge of alternatives makes it impossible to keep track of what the most common systems are, from the early academic outputs Caffe~\cite{jia2014caffe} and Keras to the vast industry-backed PyTorch~\cite{paszke2017automatic} and TensorFlow~\cite{tensorflow2015-whitepaper}.
While each of these frameworks have their own model structure and format, having a universal open-format model seemed inevitable. Microsoft and Facebook teamed up to design and implement the ONNX project~\cite{onnx}, which is an open format for interoperability among deep learning models.

Such frameworks help the models that are trained, tested and validated, to be deployed on different devices and platforms; From large models on high-end powerful devices to running smaller models on low-power tiny mobile and wearable devices. State-of-the-art system-on-chips (SoC) used as heart of mobile devices integrate heterogeneous computing elements such as multicore CPUs, GPUs, and multiple accelerators, all in a single chip~\cite{Xavier,g90}. However, this variety of computing units pose a huge threat to the deep learning frameworks since still many mobile platforms and hardware units are not fully or partially supported in these frameworks.  

This is where the compatibility issue of the models and the frameworks arises. The frameworks fully supporting the model architecture lack support for specific hardware platforms and those that support the hardware usually do not comply with the trained model format.
With the emerging demand and development of mobile technology with their radical design features~\cite{ignatov2019ai}, we observe this issue even further. In this paper, we aim at describing the challenges and obstacles that we are facing in industry when targeting variety of hardware platforms for deploying our deep learning models. Our contributions are as follows:
\begin{itemize}
    \item We explain the existing incompatibility issue between already trained models and their designed model architectures when needed to be deployed on a variety of mobile platforms or specialized hardware.
    \item We present a summary of our extensive analysis of mostly-used deep learning frameworks and the architectures they support.
    \item We categorize the main scenarios in which incompatibility occurs and deep learning model conversion is the only option. We present and discuss the solutions that we have employed in practice and provide several guidelines, based on our experience with variety of models, to overcome with the obstacles.
\end{itemize}

The rest of the paper is organized as follows: Section~\ref{sec:training} presents more detail on model training and using deep learning frameworks. In Section~\ref{sec:challenges}, we discuss in more detail the existing challenges and the deep learning frameworks. Section~\ref{sec:optimizations} introduces our practical solutions and guidelines providing illustrative examples. Finally, Section~\ref{sec:conclusions} concludes the paper.

\section{Training and Deploying Deep Learning Models}
\label{sec:training}

In this section, we first briefly present the development and training process of deep learning models using deep learning frameworks and then, we discuss how the trained models can be optimized and executed on the target hardware platform.

The very first step in the development of deep learning models is the architectural design of the model. Deep learning models are structurally designed by experts based on the user's demands and the problem. It is common to perform design space exploration and fine-tune the model to achieve the most suitable architecture before and during the training process. After this step, users employ a deep learning framework to implement their design and train the model assuming that the dataset for training the model is previously prepared and ready to use. The choice of a suitable framework depends on lots of factors such as simplicity and high-level API, flexibility, to name a few.
It is common that the target hardware platform is not clearly known during the training process. For this reason, deep learning models build on top of low-level hardware-dependent libraries providing flexibility to deploy both training and inference of the model on top of variety of hardware architectures.

The deep learning frameworks offer both training and inference processes.
Working with data, building models, optimizing model parameters, and storing trained models are all part of most deep learning workflows. 
Deep learning frameworks are usually optimised for high performance on well-known hardware platforms such as NVIDIA GPUs. They provide a clear and concise way of defining machine learning pipelines and they are user-friendly and codes are easy to understand by the users. 

After training and fine-tuning the model architecture, it can be deployed on the target devices such as CPU or GPU. They also provide highly-optimized implementation to highly utilize the hardware resources considering both model and hardware parameters.

Apart from the optimized implementation of the deep learning frameworks, the method of modifying hyperparameters in order to optimize the cost function using one of the optimization techniques is known as machine learning optimization. The cost function must be minimized since it represents the difference between the real value of an expected parameter and what the model predicts. Models can be further optimized after the training process as well. 

There are several important factors that are used to measure the deep learning inference capability:
\begin{itemize}
    \item \textbf{Latency}: Inference execution time is commonly measured in milliseconds. Low latency is essential for providing real-time inference-based services that are increasing.
    \item \textbf{Throughput}: The amount of operations performed in a given time period. Throughput, which is often calculated in inferences/second or samples/second or just operations per seconds (such as the number of floating-point operations per second, FLOPS), is important for cost-effectiveness. \item \textbf{Accuracy}: The capability of a trained neural network to deliver the correct response. This crucial metric for image classification-based applications is expressed as a top-5 or top-1 percentage.
    \item \textbf{Efficiency}: By means of efficiency, performance/watt refers to the amount of throughput delivered per unit of power. Since mobile devices, servers, and even data centers must run under set power budgets, efficiency is another important factor in cost-effective data center scaling.
    \item \textbf{Memory usage}: The amount of host and device memory that must be allocated for network inference is directly dependent on the model. This factor limits the types of models that can operate on a given inference framework. This factor is key in systems that are limited-resources.
\end{itemize}

In this line, model optimization is a key process that are performed to make models even more efficient. TensorRT~\cite{tensorrt} optimizes the network by merging layers and optimizing kernel selection. If the program specifies, the network can also be optimized to run in lower precision, such as in FP16 or int8, improving performance, in terms of execution time, and lowering memory requirements.

The optimization process, in general, is performed considering the target hardware platform, its capacity and features. Furthermore, some model optimizers such as NVIDIA TensorRT, only target specific hardware platforms. 
Despite having incompatibility to use specific model optimizers, the trained model may be still executable on the considered hardware. In this paper, we only discuss about models that are not compatible with the desired hardware regardless of the optimization process.

\section{Existing Challenges and Obstacles}
\label{sec:challenges}

In this section, we discuss the existing challenges and obstacles that we are facing when deploying some of the deep learning models on specific hardware platforms.

\begin{table*}[t!]
\centering
\begin{tabular}{  l   c  c  c  c  c  c  c  c c c c}
 \hline
  & \textbf{CPU-x86} & \textbf{CPU-Arm} & \textbf{GPU-CUDA} & \textbf{GPU-OpenCL} & \textbf{APU} & \textbf{NVDLA} & \textbf{NPU}  & \textbf{Specific ASICs}  & \textbf{FPGA}\\ 
  \hline
  TensorFlow & \Checkmark &   & \Checkmark  & \Checkmark   &   &   &   &   \\
  \hline
  TensorFlow Lite & \Checkmark & \Checkmark  & \Checkmark  &  \Checkmark &   &   &   & \Checkmark\\
  \hline
  PyTorch & \Checkmark & \Checkmark  & \Checkmark  & \Checkmark   &   &   &   & \Checkmark\\
  \hline
  PyTorch Mobile &  \Checkmark & \Checkmark  & \Checkmark  & \Checkmark   &   &   &   & \Checkmark\\
  \hline
  Tencent TNN & \Checkmark  & \Checkmark   &   & \Checkmark   &   &   & \Checkmark   & \\
  \hline
  MiniDNN~\cite{minidnn} &  & \Checkmark  &   & \Checkmark  & \Checkmark  &   &   & \\
  \hline
  Caffe & \Checkmark & \Checkmark  & \Checkmark  & \Checkmark  &   &   &   & \Checkmark & \Checkmark\\
  \hline
  Apple CoreML &  & \Checkmark  &   &   &   &   &   & \Checkmark \\
  \hline
  TensorRT & \Checkmark & \Checkmark  &  \Checkmark &   &   &   \Checkmark &  & \\
  \hline
  Apache TVM & \Checkmark & \Checkmark  & \Checkmark  & \Checkmark  &   &   &   & \Checkmark & \Checkmark \\
  \hline
  Xiaomi MACE & \Checkmark & \Checkmark  &  & \Checkmark  &   \Checkmark &   &   &  &  \\
  \hline \\
\end{tabular}
\caption{Deep learning frameworks and the hardware platforms each of them support.}
\label{tab:framework-hardware}
\end{table*} 

\subsection{Limitations in Deep Learning Frameworks} \label{ssec:limits}
We have performed an extensive analysis on the state-of-the-art and open-source deep learning frameworks in terms of supporting the commonly-used hardware platforms. Table~\ref{tab:framework-hardware} shows the frameworks that are studied and different hardware platforms that each of the frameworks supports. We studied the commonly-used frameworks with different features and specifications. As shown in Table~\ref{tab:framework-hardware}, most of the frameworks support both x86 and Arm architectures. Since most of the mobile devices employ Arm-based processors, deep learning frameworks that are developed for mobiles devices (e.g., TensorFlow Lite, PyTorch Mobile) fully support Arm-based architectures.
Regarding GPU, while most of the frameworks support NVIDIA GPUs, some of them provide limited support for AMD GPUs. Unlike CPUs and GPUs, the support for Specific ASICs (e.g., Apple Bionic, Google TPU) is very limited. Similarly, support for other deep learning accelerators such as APU, NVIDIA deep learning accelerator, and Samsung NPU are quite restricted. Note that in several of these frameworks multiple operations are not supported, such as \textit{Cast} and \textit{Dropout} operations in Tencent TNN~\cite{tencent}. 
{\color{black}
In the case of mobile devices, we have the following limitations:
\begin{enumerate}
    \item \textit{Supporting all the hardware elements in the same device}. Most of the mobile devices are based on Arm architecture, only a few of the aforementioned frameworks support the Arm-based CPU, the OpenCl-based GPU, and at the same time, the AI accelerators such as APU or NPU.
    \item \textit{Software limitations}. In frameworks such as TensorFlow Lite, PyTorch Mobile, and Apple CoreML, the models are always encapsulated inside an application which add an additional degree of complexity. 
\end{enumerate}
}

In this paper, we discuss these limitations in more detail and provide more detail regarding each of them, practical solutions that we have employed in our products as well as illustrative examples.

{\color{black}
\subsection{Emerging Development and Release of New Hardware Platforms}
Rapid development and release of new hardware platforms from numerous vendors with new features is another existing challenge we are facing in industry. The process of adapting the existing software, frameworks and libraries to support a new hardware architecture can take several months to complete. Furthermore, the early compatible software and frameworks usually lack many features and they are limited to specific items. The process of converting deep learning models can be extremely beneficial to make maximum use of the limited support.
}

\subsection{Increasing the Utilization of All Computing Elements}
As discussed earlier, latest platforms and mobile devices are integrating multiple computing elements. Furthermore, there are increasing number of applications which are required to be run concurrently. 
Therefore, it is crucial to employ and efficiently map different workloads on each of the hardware elements.

\section{Practical Solutions and Guidelines}
\label{sec:optimizations}

In this section, we present different challenges we faced during model conversion process and we have proposed and present solutions for each case. We demonstrate the process and the modifications using examples we faced when working with state-of-the-art models.

\subsection{Hardware Platform Support}

As far as both the target hardware platform and the deep learning model format are supported by the deep learning framework, the model can be fully deployed.
However, as discussed in Section~\ref{ssec:limits}, first limitation, the deep learning framework may not support some of the existing hardware platforms or for specific computing units of them. For instance, Android or iOS devices require specific deep learning frameworks to support the operating system and its hardware. Furthermore, hardware processors are developed based on Arm, x86, or other instruction set architectures (ISA) which require compliant software accordingly. On the other hand, latest devices integrate variety of computation units such as CPU, Graphics Processing Unit (GPU), Digital Signal Processor (DSP), and specialized hardware accelerators such as the AI processing unit (APU) in MediaTek P90 and G90~\cite{g90}, the deep learning accelerator in NVIDIA Xavier~\cite{Xavier}, Samsung NPU, and the Qualcomm Hexagon processor~\cite{codrescu2014hexagon,hexagon}. The user usually needs to run a model on specific compute unit, e.g., the APU, which may not be supported by the baseline deep learning framework. 

Providing support for new hardware platforms require enormous engineering and implementation effort as well as sufficient knowledge on both the deep learning framework and the target hardware architecture. Therefore, this option is not usually the choice of many developers. Note that despite the complexity of this process, it is only feasible if the framework is open-source and sufficient details regarding the hardware architecture is available and not all the vendors are revealing this level of detailed information publicly.

In practice, for hardware platforms that are not supported by the baseline deep learning framework, the conversion of the baseline model format to a universal format and, subsequently, to another framework can be assessed. 
After extensive analysis and considering the models presented in Table~\ref{tab:framework-hardware}, we use Tencent TNN, Xiaomi MACE, and MiniDNN frameworks which are evolving along with the evolution of the mobile hardware market.
To use these frameworks, the baseline trained model needs to be converted into a universal format, such as converting from PyTorch to ONNX~\cite{onnx}. 
This requires that the converted model is supported by the TNN, MACE, and MiniDNN. We discuss in more detail how we deal with unsupported models later in this section. 
Despite these complexities in this process, in most cases, the format conversion can be done smoothly, thanks to the variety of model conversion tools.
Given that the conversion to the target framework is done, TNN-like frameworks are proving to be ideal to support the target mobile hardware platforms.

It should be mentioned that in several cases, the baseline model is developed elsewhere without taking all the possible target hardware into consideration. Also, in industry, due to dataset confidentiality, the training process is only performed by the dataset owner or by accessing and using their data centers. This leads us to the second limitation as discussed in Section~\ref{ssec:limits} since it brings model incompatibility issues.

\subsection{Software Limitations and Proposed Solutions}
Overall, the software limitations arise from the choice of framework:
\begin{itemize}
    \item First, the use of sophisticated frameworks such as PyTorch Mobile or TensorFlow Lite enforces the use of Android Studio to create the Android Application Package (APK) for Android devices. The APK includes the model required to run the inference on the device. This adds an additional layer of drawbacks that are derived from Android Studio itself. A practical example that we faced with our products is that the model input resolution is strictly limited to what Android Studio supports.
    \item Second, recent frameworks such as TNN and the like might have issues supporting all the state-of-the-art layers and operations. However, they support a bare-metal deployment of models to run inference on mobile devices. 

\end{itemize}

Since issues such as limited resolution is not an option for several domains such as autonomous driving, the natural choice is to proceed with the use of recent frameworks.

To use these recent frameworks for the conversion process, tools like \texttt{torch.onnx.export} are highly helpful, however, we still face one major issue:
The converted model is not compatible with the new framework which supports the target hardware. The baseline model includes new user-defined custom layer which cannot be supported by conversion tools as well as other formats.
In case the model conversion is not feasible, there is no choice but to modify the baseline model and generate a new model accordingly. The new model or a converted format of it needs to be compatible with the deep learning framework that is supporting the target hardware.

We have categorized different scenarios in which unsupported layers needed to be transformed into compatible and supported layers (operations) by the new framework. These are the following four main categories: \circled{1} Modifications on the baseline model, \circled{2} Modifying the model architecture and preforming a full retraining process, \circled{3} Performing some operations as a post-processing stage after the deployment of the new model format, and \circled{4} Implementing custom layers in the new deep learning framework. In the following, we discuss these categories in more detail providing illustrative examples.

\begin{figure}[t!]
    \centering
    \includegraphics[width=0.9\columnwidth]{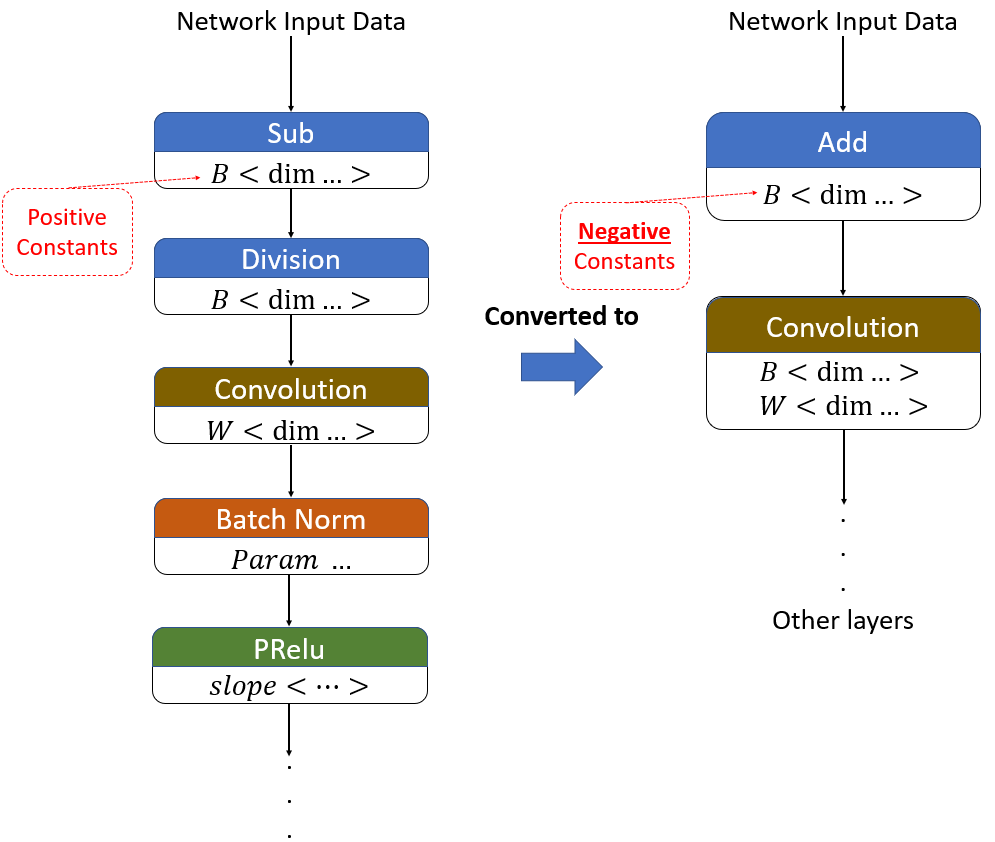}
    \caption{An example of the first scenario with applied modifications on the baseline model.}
    \label{fig:scenario-1}
    \vspace{-0.4cm}
\end{figure}

\textbf{\circled{1} Modifications on the Baseline Model}.
The first experienced scenario is to have the unsupported layer among the very first or the last layers of the model. For some layers performing specific operation, it is possible to use a similar and supported layer in addition to applying constant values in such a way that the combined operation mimics exactly the same as the baseline layer. For instance, as shown in Figure~\ref{fig:scenario-1}, we have a model with the \textit{Subtract} layer which is not supported and needs to be replaced with one or several layers to do the same operation. In this example, we added a \textit{Add} layer with negative constant values mimicking the exact same ``substract'' operation. In this example, the modifications are performed with no need to retraining the model.

\begin{figure}[t!]
    \centering
    \includegraphics[width=0.9\columnwidth]{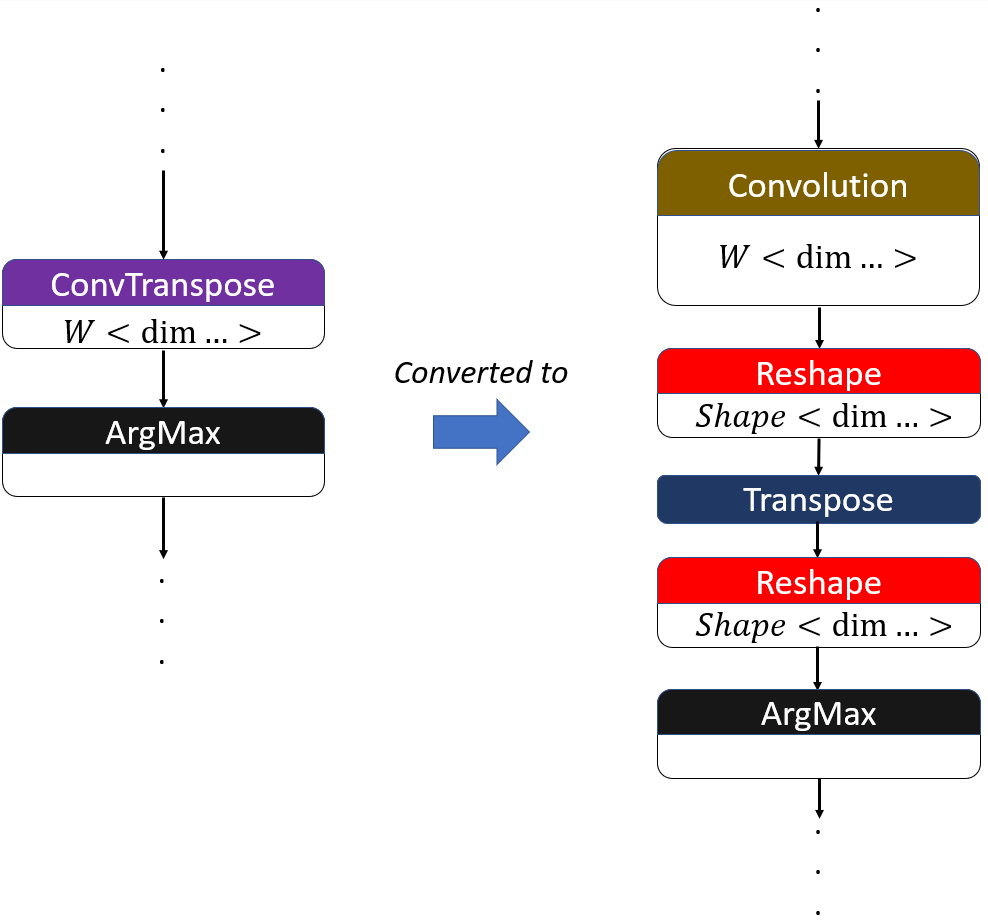}
    \caption{An example of the second scenario with an incompatible layer, applied changes requiring retraining the model.}
    \label{fig:scenario-2}
    \vspace{-0.5cm}
\end{figure}

\textbf{\circled{2} Modifying the Model Architecture and Performing Full Retraining}.
In this scenario, we face an unsupported layer which requires to be replaced with one or several layers, however, it needs a model retraining process to train and fine-tune the parameters of the new layers. As an illustrative example shown in Figure~\ref{fig:scenario-2}, the target deep learning framework does not support the \textit{ConvTranspose} (deconvolution) layer. In this case, a series of operations in several new layers are able to imitate the deconvolution process. In this case a ``Convolution" layer followed by two ``Reshape" and one ``Transpose" layers are replacing the ConvTranspose layer and the layers before and the rest of the network remain the same as in the baseline model architecture. To train the parameters of these new layers, the model needs to go through the training process again. Based on our experiments, after retraining the model in such a scenario, we have observed very similar accuracy as in the baseline model. Since the model structure is changed, a negligible change in the accuracy of the model is inevitable.

\begin{figure}[t!]
    \centering
    \includegraphics[width=0.9\columnwidth]{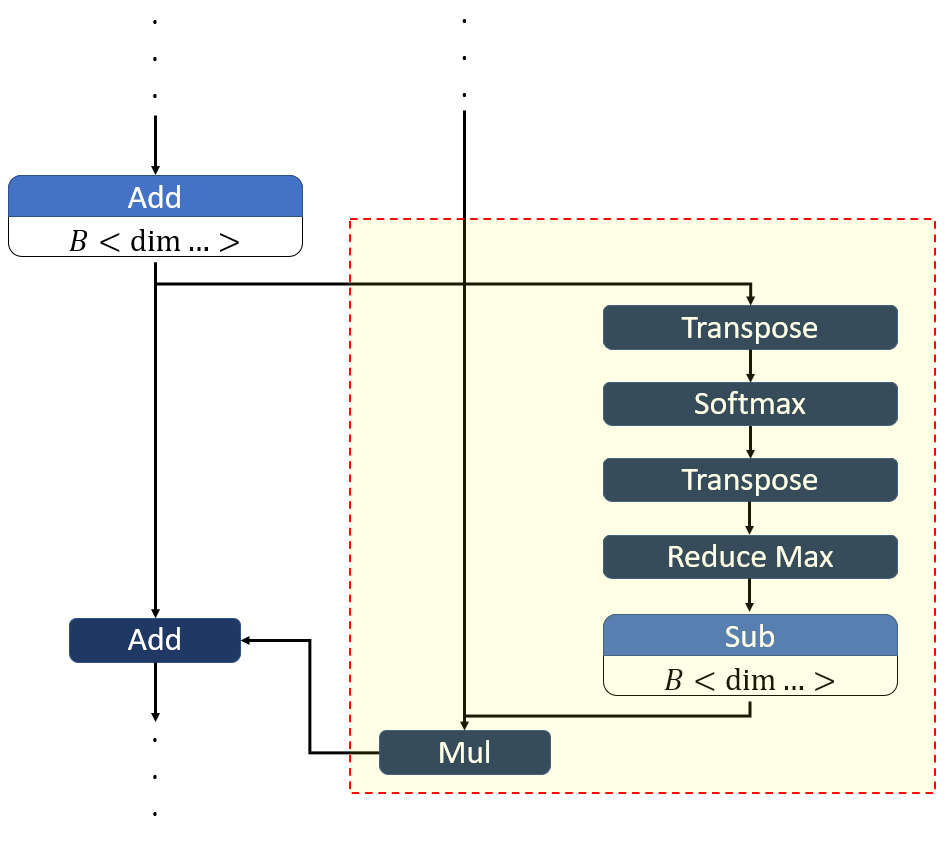}
    \caption{An example of third scenario in which some incompatible layers are removed and then performed as a post-processing step.}
    \label{fig:scenario-3}
    \vspace{-0.5cm}
\end{figure}

\textbf{\circled{3} Post-processing Stage after Model Deployment}.
When the unsupported layer is among the trailing layers of the model, in some cases, it is feasible to perform those operations out of the model and as post-processing steps. Therefore, in such a scenario, we can remove the layer(s) from the model without the need to retrain the model and adapt the parameters. In the example illustrated in Figure~\ref{fig:scenario-3}, despite the \textit{Subtract} layer, the \textit{Softmax} layer is also not supported in the target deep learning framework. In order to eliminate this issue, the entire sequence of the highlighted layers can be removed from the model and performed as a post-processing step. 
{\color{black}
In this case, we need multiple nodes to be available as the output of the model graph especially when there are skip connections such as inputs to the \textit{Mul} node in Figure~\ref{fig:scenario-3}. Many deep learning and inference frameworks support only one output node. In such cases, all of the multiple nodes have to be combined into one in a separable way to be available for post-processing. 
}

\textbf{\circled{4} Implementing Custom Layers in the Deep Learning Framework}.
In some cases, it is not trivial to perform the actions in the aforementioned scenarios or they might not be cost-effective. Most of the deep learning models offer the possibility of implementing custom user-defined layers. Note that some layers may not be straightforward to implement or an efficient and optimized implementation may not be trivial. Upon the possibility of implementing custom layers, there is a trade-off between this option or the modification of the model as discussed earlier.
{\color{black}
Note that when converting models from one deep learning framework to another for running inference on mobile devices, it is required to implement these custom layers in both the baseline framework as well as in the framework that is running on the mobile device. For instance, for the miniDNN to run inference of a PyTorch model, the model first needs to be converted to Caffe format and then, to miniDNN. Support for layers need to be established on both Caffe and miniDNN frameworks resulting in twofold effort. 
}

We would like to emphasize that our experience illustrated from the above four typical scenarios can highly fill the gap between research and deployment of the new layers on various hardware platforms.
\section{Conclusions}
\label{sec:conclusions}

In this paper, we first discussed the current barriers and roadblocks to deploying deep learning models on mobile and edge devices. 
Second, by studying various deep learning frameworks, we showed a subset of them are able to map our models to the desired hardware for inference. 
Third, we also provided solutions to tackle layer incompatibility issues by suggesting alternative layers, retraining schemes, and post-processing split which are typical problems in model conversion. We believe that the raised problems and the proposed solutions can pave the way towards deploying variety of deep learning models on specific and newly developed hardware platforms.

\bibliographystyle{IEEEtran}
\bibliography{bibliography}

\end{document}